\documentclass[10pt,twocolumn,letterpaper]{article}

\usepackage{cvpr}
\usepackage{times}
\usepackage{epsfig}
\usepackage{graphicx}
\usepackage{amsmath}
\usepackage{amssymb}
\usepackage{multirow}
\usepackage{rotating}
\usepackage{color}
\usepackage[skip=0pt,font=footnotesize]{caption}
\usepackage{subcaption}
\usepackage{float}

\newcommand \hh {\mathbf{h}}
\newcommand \oo {\mathbf{o}}
\newcommand \xx {\mathbf{x}}
\newcommand \zero {\textbf{0}}

 \cvprfinalcopy 

\def\cvprPaperID{1981} 

\ifcvprfinal\fi
\begin{document}

\title{DAG-Recurrent Neural Networks For Scene Labeling}

\author{Bing Shuai{$^*$}, Zhen Zuo{$^*$}, Gang Wang, Bing Wang\\
School of Electrical and Electronic Engineering, Nanyang Technological University, Singapore.\\
{\tt\small \{bshuai001,wanggang,zzuo1,wang0775\}@ntu.edu.sg}
\thanks{Equal Contribution}}


\makeatletter
\g@addto@macro\@maketitle{
  \begin{figure}[H]
  \setlength{\linewidth}{\textwidth}
  \setlength{\hsize}{\textwidth}

  \newcommand{\InsertImage}[2]{
  \begin{subfigure}[t]{0.12\textwidth}
  \centering
  \includegraphics[width=\textwidth]{#1}
  \caption*{\scriptsize{{#2}}}
  \end{subfigure}
  \hspace{-8pt}
  }
  \begin{center}
  \InsertImage{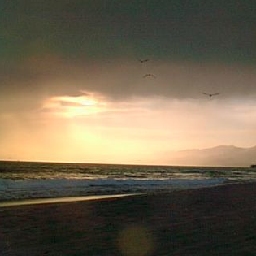}{Input Image}
  \InsertImage{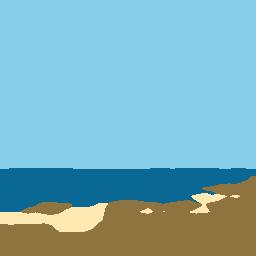}{CNN}
  \InsertImage{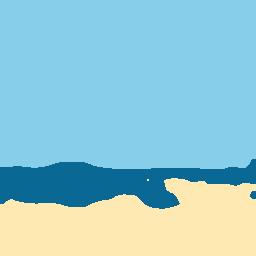}{DAG-RNN}
  \InsertImage{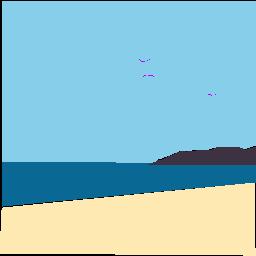}{Ground Truth}
  \hspace{5pt}
  \InsertImage{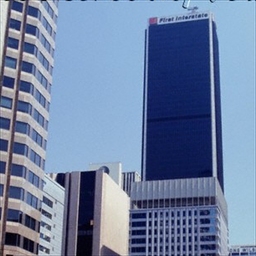}{Input Image}
  \InsertImage{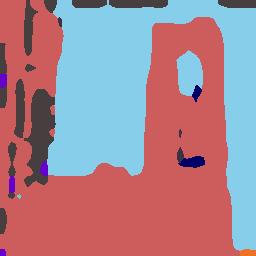}{CNN}
  \InsertImage{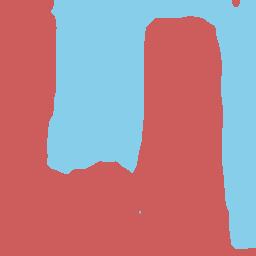}{DAG-RNN}
  \InsertImage{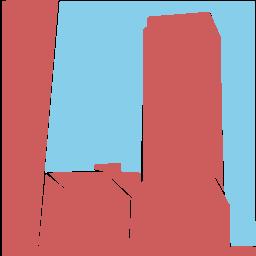}{Ground Truth}\\
  \end{center}
  \caption{ With the local representations extracted from Convolutional Neural Networks (CNNs), the `sand' pixels (in the first image) are likely to be misclassified as `road', and the `building' pixels (in the second image) are easy to get confused with `streetlight'. Our DAG-RNN is able to significantly boost the discriminative power of local representations by modeling their contextual dependencies. As a result, it can produce smoother and more semantically meaningful labeling map. The figure is best viewed in color. }
  \label{Figure:intro}
  \end{figure}
}
\makeatother

\maketitle

\begin{abstract}
  In image labeling, local representations for image units are usually generated from their surrounding image patches, thus long-range contextual information is not effectively encoded. In this paper, we introduce recurrent neural networks (RNNs) to address this issue.
  Specifically, directed acyclic graph RNNs (DAG-RNNs) are proposed to process DAG-structured images, which enables the network to model long-range semantic dependencies among image units.
  Our DAG-RNNs are capable of tremendously enhancing the discriminative power of local representations, which significantly benefits the local classification. Meanwhile, we propose a novel class weighting function that attends to rare classes, which phenomenally boosts the recognition accuracy for non-frequent classes. Integrating with convolution and deconvolution layers, our DAG-RNNs achieve new state-of-the-art results on the challenging SiftFlow, CamVid and Barcelona benchmarks.
\end{abstract}

\section{Introduction}

Scene labeling refers to associating one of the semantic classes to each pixel in a scene image. It is usually defined as a multi-class classification problem based on their surrounding image patches. However, some classes may be indistinguishable in a close-up view. As an example in Figure \ref{Figure:intro}, the `sand' and `road' pixels are hard to be distinguished even for humans with limited context. In contrast, their differentiation becomes conspicuous when they are considered in the global scene. Thus, how to equip local features with a broader view of contextual awareness is a pivotal issue in image labeling.

In this paper, recurrent neural networks (RNNs) \cite{graves2013speech}\cite{pascanu2012difficulty} are introduced to address this issue by modeling the contextual dependencies of local features. 
Specifically, we adopt undirected cyclic graphs (UCG) to model the interactions among image units.
Due to the loopy property of UCGs, RNNs are not directly applicable to UCG-structured images. Thus, we decompose the UCG to several directed acyclic graphs (DAGs, and four DAGs are used in our experiments).
In other words, an UCG-structured image is approximated by the combination of several DAG-structured images. Then, we develop the DAG-RNNs, a generalization of RNNs \cite{graves2012offline}\cite{graves2013speech}, to process DAG-structured images. Each hidden layer is generated independently through applying DAG-RNNs to the corresponding DAG-structured image, and they are integrated to produce the context-aware feature maps. In this case, the local representations are able to embed the abstract gist of the image, so their discriminative power are enhanced remarkably.

We integrate the DAG-RNNs with the convolution and deconvolution layers, thus giving rise to an end-to-end trainable full labeling network.
Functionally, the convolution layer transforms RGB raw pixels to compact and discriminative representations. Based on them, the proposed DAG-RNNs model the contextual dependencies of local features, and output the improved context-aware representation. The deconvolution layer upsamples the feature maps to match the dimensionality of the desired outputs. Overall, the full labeling network accepts variable-size images  and generates the corresponding dense label prediction maps in a single feed-forward network pass.
Furthermore, considering that the class frequency distribution is highly imbalanced in natural scene images, we propose a novel class weighting function that attends to rare classes.

We test the proposed labeling network on three popular and challenging scene labeling benchmarks (SiftFlow \cite{liu2009nonparametric}, CamVid \cite{BrostowSFC:ECCV08} and Barcelona\cite{tighe2010superparsing}). On these datasets,  we show that our DAG-RNNs are capable of greatly enhancing the discriminative power of local representations, which leads to dramatic performance improvements over baselines (CNNs, even the VGG-verydeep-16 network \cite{simonyan2014very}). Meanwhile, the proposed class weighting function is able to boost the recognition accuracy for rare classes.
Most importantly, our full labeling network significantly outperforms current state-of-the-art methods.

Next, related work are firstly reviewed, compared and discussed in Section \ref{Section:related_work}. Section \ref{Section:approach} elaborates the details of the DAG-RNNs and how they are applied to image labeling. Besides, it presents the details of the full labeling network and the class weighting function. The detailed experimental results and analysis are presented in Section \ref{Section:experiments}. In the end, section \ref{Section:conclusion} concludes the paper.

\section{Related Work}
Scene labeling (also termed as scene parsing, semantic segmentation) is one of the most challenging problems in computer vision. It has attracted more and more attention in recent years. Here we would like to highlight and discuss three lines of works that are most relevant to ours.

The first line of work is to explore the contextual modeling. One attempt is to encode context into local representation. For example, Farabet et al. \cite{farabet2013learning} stacks surrounding contextual windows from different scales; Pinheiro et al. \cite{pinheiro2014recurrent} increases the size of input windows. Sharma et al. \cite{sharma2014recursive} adopts recursive neural networks to propagate global context to local regions.
However, they do not consider any structure for image units, thus their correlations are not effectively captured. In contrast, we interpret the image as an UCG, within which the connections allow the DAG-RNNs to explicitly model the dependencies among image units.
Another attempt is to pass context to local classifiers by building probabilistic graphical models (PGM). For example, Shotton et al. \cite{shotton2006textonboost} formulates the unary and pairwise features in a 2nd-order Conditional Random Field (CRF). Zhang et al.\cite{zhang2012efficient} and Roy et al. \cite{royscene} build a fully connected graph to enforce higher order labeling coherence. Shuai et al.\cite{shuai2015integrating} models the global-order dependencies in a non-parametric framework to disambiguate the local confusions. Our work also differs from them. First, the label dependencies are defined in terms of compatibility functions in PGM, while such dependencies are modeled through a recurrent weight matrix in RNNs. Moreover, the inference of PGM is inefficient as the convergence of local beliefs usually takes many iterations. In contrast, RNNs only need a single forward pass to propagate the local information.

Some of the previous work exploit `recurrent' ideas in a different way. They generally refer to applying the identical model recurrently at different iterations (layers). For example, Pinheiro et al.\cite{pinheiro2014recurrent} attachs the RGB raw data with the output of the Convolutional Neural Network (CNN) to produce the input for the same CNN in the next layer. Tu et al.\cite{tu2008auto} augments the patch feature with the output of the classifier to be the input for the next iteration, and the classifier parameters are shared across different iterations. Zheng et al.\cite{zheng2015conditional} transforms Conditional Random Fields (CRF) to a neural network, so the inference of CRF equals to applying the same neural network recurrently until some fixed point (convergence) is reached. Our work differs from them significantly. They model the context in the form of intermediate outputs (usually local beliefs), which implicitly encodes the neighborhood information. In contrast, the contextual dependencies are modeled explicitly in DAG-RNNs by propagating information via the recurrent connections.

Recurrent neural networks (RNNs) have achieved great success in temporal dependency modeling for chain-structured data, such as natural language and speeches.
Zuo et al. \cite{zuo2015convolutional} applies 1D-RNN to model weak contextual dependencies in image classification.
Graves et al. \cite{graves2012offline} generalizes 1D-RNN to multi-dimensional RNN (MDRNN) and applies it to offline arabic handwriting recognition. Shuai et al.\cite{shuai2015quaddirectional} also adopts 2D-RNN to real-world image labeling. Recently, Tai et al.\cite{tai2015improved} and Zhu et al.\cite{zhu2015long} demonstrate that considering tree structure (constituent / parsing trees for sentences) is beneficial for modeling the global representation of sentences. Our proposed DAG-RNN is a generalization of chain-RNNs \cite{bengio2013advances}\cite{irsoy2014opinion}, tree-RNNs \cite{tai2015improved}\cite{zhu2015long} and 2D-RNNs \cite{graves2012offline}\cite{shuai2015quaddirectional}, and  it enables the network to model long-range semantic dependencies for graphical structured images. The most relevant work to ours is \cite{shuai2015quaddirectional}. In comparison with which, (1), we generalize 2D-RNN to DAG-RNN and show benefits in quantitative labeling performance; (2), we integrate the convolution layer, deconvolution layer with our DAG-RNNs to a full labeling network; and (3), we adopt a novel class weighting function to address the extremely imbalanced class distribution issue in natural scene images. To the best of our knowledge, our work is the first attempt to integrate the convolution layers with RNNs in an end-to-end trainable network for real-world image labeling.
Moreover, the proposed full network achieves state-of-the-art on a variety of scene labeling benchmarks.



\label{Section:related_work}
\section{Approach}
\label{Section:approach}

To densely label an image $I$, the image is processed by three different functional layers sequentially: (1), Convolution layer produces the corresponding feature map $\xx$. Each feature vector in $\xx$ summarizes the information from a local region in $I$. (2), DAG-RNNs model the contextual dependency among elements in $\xx$, and generates the intermediate feature map $\hh$, whose element is a feature vector that implicitly embeds the abstract gist of the image. (3), Deconvolution layer \cite{long2015fully} upsamples the feature maps. From which, the dense label prediction maps are derived. We start by introducing the proposed DAG-RNNs, and the details of the full network are elaborated in the following sections.

\subsection{RNNs Revisited}
\label{Section:1d_rnn}
A recurrent neural network (RNN) is a class of artificial neural network that has recurrent connections, which equip the network with memory.
In this paper, we focus on the Elman-type network \cite{elman1990finding}.
Specifically, the hidden layer $h^{(t)}$ in RNNs at time step $t$ is expressed as a non-linear function over current input $x^{(t)}$ and hidden layer at previous time step $h^{(t-1)}$. The output layer $y^{(t)}$ is connected to the hidden layer $h^{(t)}$.

Mathematically, given a sequence of inputs $\{ x^{(t)} \}_{t=1:T}$, an Elman-type RNN operates by computing the following hidden and output sequences:
\begin{equation}
\begin{aligned}
 h^{(t)} &= f(Ux^{(t)} + Wh^{(t-1)} + b) \\
 y^{(t)} &= g(Vh^{(t)} + c)
\end{aligned}
\label{Equation:elman_rnn}
\end{equation}

where $U, W$ are weight matrices between the input and hidden layers, and among the hidden units themselves, while $V$ is the output matrix connecting the hidden and output layers; $b, c$ are corresponding bias vectors and $f(\cdot), g(\cdot)$ are element-wise nonlinear activation functions. The initial hidden unit $h^{(0)}$ is usually assumed to be $\textbf{0}$. The local information $x^{(t)}$ is progressively stored in the hidden layers by applying Equation \ref{Equation:elman_rnn}. In other words, the contextual information (the summarization of past sequence information) is explicitly encoded into local representation $h^{(t)}$, which improves their representative power dramatically in practice.

Training a RNN can be achieved by optimizing a discriminative objective with a gradient-based method. Back Propagation through time (BPTT) \cite{werbos1990backpropagation} is usually used to calculate the gradients. This method is equivalent to unfolding the network in time and using back propagation in a very deep feed-forward network except that the weights across different time steps (layers) are shared.

\subsection{DAG-RNNs}
\label{Section:dag_rnn}


\label{Section:quaddirectionality}
\begin{figure}[t]
\centering
\includegraphics[width=0.45\textwidth]{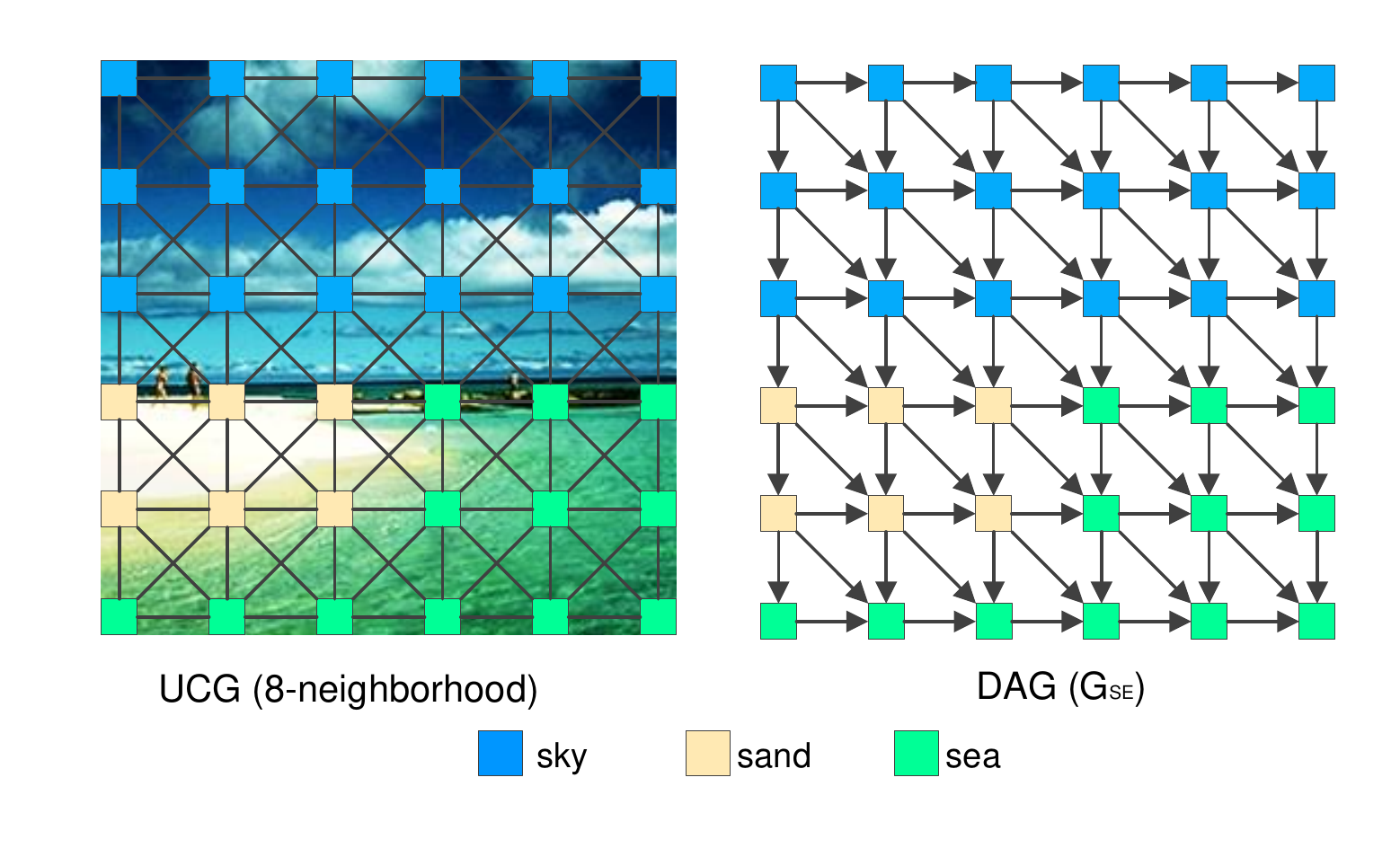}
\caption{An 8-neighborhood UCG and one of its induced DAG in the southeastern (SE) direction.}
\label{Figure:ucg_dag}
\end{figure}

The aforementioned RNN is designed for chain-structured data (e.g. sentences or speeches), where temporal dependency is modeled. However, interactions among image units are beyond chain. In other words, traditional chain-structured RNNs are not suitable for images. Specifically, we can reshape the feature tensor $\xx \in \mathbb{R}^{h \times w \times d}$ to $\hat{\xx} \in \mathbb{R}^{(h\cdot w) \times d}$, and generate the chain representation by connecting contiguous elements in $\hat{\xx}$. Such a structure loses spatial relationship of image units, as two adjacent units in image plane may not necessarily be neighbors in the chain. The graphical representations that respect the 2-D neighborhood system are more plausible solutions, and they are pervasively adopted in probabilistic graphical models (PGM). Therefore in this work, undirected cyclic graphs (UCG , an example is shown in Figure \ref{Figure:ucg_dag}) are used to model the interactions among image units.



Due to the loopy structure of UCGs, they are unable to be unrolled to an acyclic processing sequence. Therefore, RNNs are not directly applicable to UCG-structured images.
To address this issue, we approximate the topology of UCG by a combination of several directed acyclic graphs (DAGs), each of which is applicable for our proposed DAG-RNNs (one of the induced DAGs is depicted in Figure \ref{Figure:ucg_dag}). Namely, an UCG-structured image is represented as the combination of a set of DAG-structured images.  We now start introducing the detailed mechanism of our DAG-RNNs here, and later elaborate how they are applied to UCG-structured images in the next section.



We first assume that an image $I$ is represented as a DAG $\mathcal{G} = \{\mathcal{V}, \mathcal{E} \}$, where $\mathcal{V} = \{v_i\}_{i=1:N}$ is the vertex set and $\mathcal{E} = \{e_{ij}\}$ is the arc set ($e_{ij}$ denotes an arc from $v_i$ to $v_j$). The structure of the hidden layer $\hh$ follows the same topology as $\mathcal{G}$. Therefore, a forward propagation sequence can be generated by traversing $\mathcal{G}$, on the condition that one node should not be processed until all its predecessors are processed.
The hidden layer $\hh^{(v_i)}$ is represented as a nonlinear function over its local input $\xx^{(v_i)}$ and the summarization of hidden representation of its predecessors. The local input $\xx^{(v_i)}$ is obtained by aggregating (e.g. average pooling) from constituent elements in the feature tensor $\xx$.
In detail, the forward operation of DAG-RNNs is calculated by the following equations:
\begin{equation}
\begin{aligned}
\hat{\hh}^{(v_i)} & = \!\!\!\!\! \sum_{v_j \in \mathcal{P}_{\mathcal{G}}(v_i)} \!\!\!\!\!\hh^{(v_j)} \\
\hh^{(v_i)} & = f(U\xx^{(v_i)} + W\hat{\hh}^{(v_i)} + b) \\
\oo^{(v_i)} & = g(V\hh^{(v_i)} + c)
\end{aligned}
\label{Equation:dag_rnn}
\end{equation}
where $\xx^{(v_i)}, \hh^{(v_i)}, \oo^{(v_i)}$ are the representations of input, hidden and output layers located at $v_i$ respectively, $\mathcal{P}_{\mathcal{G}}(v_i)$ is the direct predecessor set of vertex $v_i$ in the graph $\mathcal{G}$,
 $\hat{\hh}^{(v_i)}$ summarizes the information of all the predecessors of $v_i$.
 Note that the recurrent weight $W$ in Equation \ref{Equation:dag_rnn} is shared across all predecessor vertexes in $\mathcal{P}_{\mathcal{G}}(v_i)$. We may learn a specific recurrent matrix $W$ for each predecessor when vertexes (except source and sink vertex) in the DAG $\mathcal{G}$ have a fixed number of predecessors.
In this case,  a finer-grained dependency may be captured.

The derivatives are computed in the backward pass, and each vertex is processed in the reverse order of forward propagation sequence. Specifically, to derive the gradients at $v_i$, we look at equations (besides Equation \ref{Equation:dag_rnn}) that involve $\hh^{(v_i)}$ in the forward pass:
\begin{equation}
\small
\begin{aligned}
&\forall v_k \in \mathcal{S}_{\mathcal{G}}(v_i)\\
&\hh^{(v_k)} = f(U\xx^{(v_k)} + W\hh^{(v_i)} + W\tilde{\hh}^{(v_k)}  + b)\\
&\tilde{\hh}^{(v_k)}  = \!\!\!\!\!\!\!\!\!\! \sum_{v_j \in \mathcal{P}_{\mathcal{G}}(v_k) - \{v_i\}}  \!\!\!\!\!\!\!\!\!\!\!\! \hh^{(v_j)}\\
\end{aligned}
\label{Equation:dag_rnn_vi}
\end{equation}
where $\mathcal{S}_{\mathcal{G}}(v_i)$ is the direct successor set for vertex $v_i$ in the graph $\mathcal{G}$. It can be inferred from Equation \ref{Equation:dag_rnn}, \ref{Equation:dag_rnn_vi} that the errors backpropagated to the hidden layer  ($d\hh^{(v_i)}$) at $v_i$ have two sources: direct errors from $v_i$ ($\frac{\partial \oo^{(v_i)}}{\partial \hh^{(v_i)}}$), and summation over indirect errors propagated from its successors ($\sum_{v_k}\!\!\frac{\partial \oo^{(v_k)}}{\partial \hh^{(v_k)}} \frac{\partial \hh^{(v_k)}}{\partial \hh^{(v_i)}}$). The derivatives at $v_i$ can then be computed by the following equations:
\footnote{To save space, we omit the expression for $\Delta b$ and $\Delta c$ here as they can be inferred trivially from Equation \ref{Equation:dag_bptt}.}

\begin{equation}
\small
\begin{aligned}
&\Delta V^{(v_i)} =  g'(\oo^{(v_i)}) (\hh^{(v_i)})^T \\
&d\hh^{(v_i)} = V^T g'(\oo^{(v_i)})+ \!\!\!\!\! \sum_{v_k \in \mathcal{S}_{\mathcal{G}}{(v_i)}} \!\!\!\!\! W^T d\hh^{(v_k)} \circ f'(\hh^{(v_k)}) \\
&\Delta W^{(v_i)} =  \!\!\!\sum_{v_k \in \mathcal{S}_{\mathcal{G}}(v_i)}\!\!\!\!\! d\hh^{(v_k)} \circ f'(\hh^{(v_k)})(\hh^{(v_i)})^T \\
&\Delta U^{(v_i)} =  d\hh^{(v_i)} \circ f'(\hh^{(v_i)}) (\xx^{(v_i)})^T
\end{aligned}
\label{Equation:dag_bptt}
\end{equation}
where $\circ$ denotes the Hadamard product, $g'(\cdot) = \frac{\partial L}{\partial \oo(\cdot)}\frac{\partial \oo(\cdot)}{\partial g}$ is the derivative of loss function $L$ with respect to the output function $g$, and $f'(\cdot) = \frac{\partial \hh}{\partial f}$.
It is  the second term of $d\hh^{(v_i)}$ in Equation \ref{Equation:dag_bptt} that enables DAG-RNNs to propagate local information, which behaves similarly to the message passing \cite{yedidia2003understanding} in probabilistic graphic models.

\subsection{Decomposition}
We decompose the UCG $\mathcal{U}$ to a set of DAGs $\mathcal{G}^{\mathcal{U}}=\{\mathcal{G}_1, \ldots, \mathcal{G}_d, \ldots \}$. Hence, the UCG-structured image is represented as the combination of a set of DAG-structured images.
Next, DAG-RNNs are applied independently to each DAG-structured image, and the corresponding hidden layer $\hh_d$ is generated.
The aggregation of the independent hidden layers yields the output layer $\oo$. These operations can be mathematically expressed as follows:
\begin{equation}
\small
\begin{aligned}
&\hh_{d}^{(v_i)} = f(U_d \xx^{(v_i)} + \!\!\!\!\! \sum_{v_j \in \mathcal{P}_{\mathcal{G}_d}(v_i)} \!\!\!\!\!\!\! W_d \hh_{d}^{(v_j)} + b_d) \\
&\oo^{(v_i)} = g(\sum_{\mathcal{G}_d \in \mathcal{G}^{\mathcal{U}} } V_d \hh_{d}^{(v_i)} + c)\\
\end{aligned}
\label{Equation:quad_dag_rnn}
\end{equation}
where $U_d, W_d, V_d$ and $b_d$ are weight matrices and bias vector for the DAG $\mathcal{G}_d$, $P_{\mathcal{G}_d}(v_i)$ is the direct predecessor set of vertex $v_i$ in $\mathcal{G}_{d}$.
This strategy is reminiscent of the tree-reweighted max-product algorithm (TRW) \cite{wainwright2005new}, which represents the problem on the loopy graphs as a convex combination of tree-structured problems.

We consider the following criterions for the decomposition.  Topologically, the combination of DAGs should be equivalent to the UCG $\mathcal{U}$, so any two vertexes can be reachable. Besides, the combination of DAGs should allow the local information to be routed to anywhere in the image.
In our experiment, we use the four context propagation directions (southeast, southwest, northwest and northeast) suggested by \cite{graves2012offline}\cite{shuai2015quaddirectional} to decompose the UCG.
One example of the induced DAG of the 8-neighborhood UCG in the southeast direction is shown in Figure \ref{Figure:ucg_dag}.

\subsection{Full Labeling Network}
\begin{figure}[t]
\begin{center}
  \includegraphics[width=0.5\textwidth]{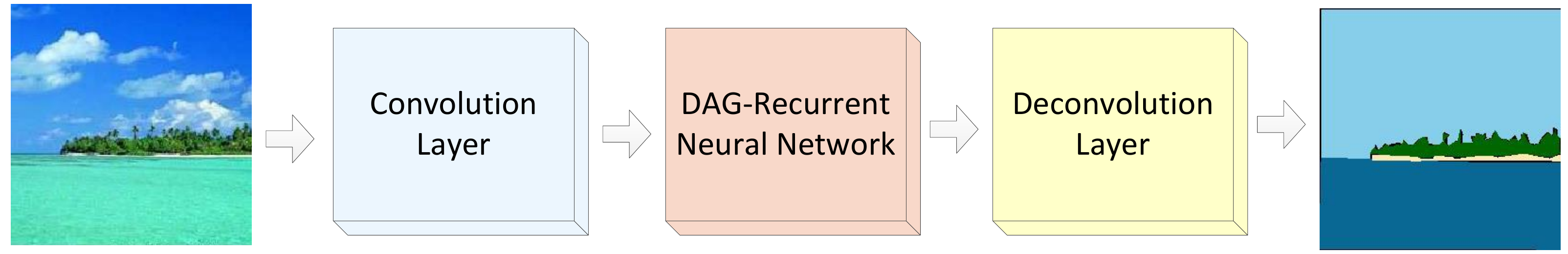}
\end{center}
\caption{The architecture of the full labeling network, which consists of three functional layers: (1), convolution layer: it produces discriminative feature maps; (2), DAG-RNN: it models the contextual dependency among elements in the feature maps; (3), deconvolution layer: it upsamples the feature maps to output the desired sizes of label prediction maps.}
\label{Figure:pipeline}
\end{figure}

The skeleton architecture of the full labeling network is illustrated in Figure \ref{Figure:pipeline}. The network is end-to-end trainable, and it takes input as raw RGB images with any size. It outputs the label prediction maps with the same size of inputs.


The convolution layer is used to produce compact yet highly discriminative features for local regions.
Next, the proposed DAG-RNN is used to model the semantic contextual dependencies of local representations.
Finally, the deconvolution layer \cite{long2015fully} is introduced to upsample the feature maps by learning a set of deconvolution filters, and it enables the full labeling network to produce the desired size of label prediction maps.

To train the network, we adopt the average weighted cross entropy loss. It is formally written as:
\begin{equation}
\small
L = -\frac{1}{N}\sum_{v_i \in I}\sum_{j=1}^{c} \mathbf{w}_j log (\oo^{(v_i)}_j  \ \mathbf{y}^{(v_i)}_j)
\end{equation}
where $N$ is the number of image units in image $I$; {$\mathbf{w}$ is the class weight vector}, in which $\mathbf{w}_j$ stands for the weight for class $j$; $\mathbf{y}^{(v_i)}$ is the binary label indicator vector for the image unit located in $v_i$, and $\oo^{(v_i)}$  stands for the corresponding class likelihood vector. The errors propagated from DAG-RNNs to the convolution layer for image unit $v_i$ are calculated based on the following equations:
\begin{equation}
\small
\Delta x^{(v_i)} =  \sum_{\mathcal{G}_d \in \mathcal{G}^{\mathcal{U}}} U_d^T d\hh_d^{(v_i)} \circ f'(\hh_d^{(v_i)})
\end{equation}

\subsection{Attention to Rare Classes}
\begin{figure}
\begin{center}
  \includegraphics[width=0.239\textwidth]{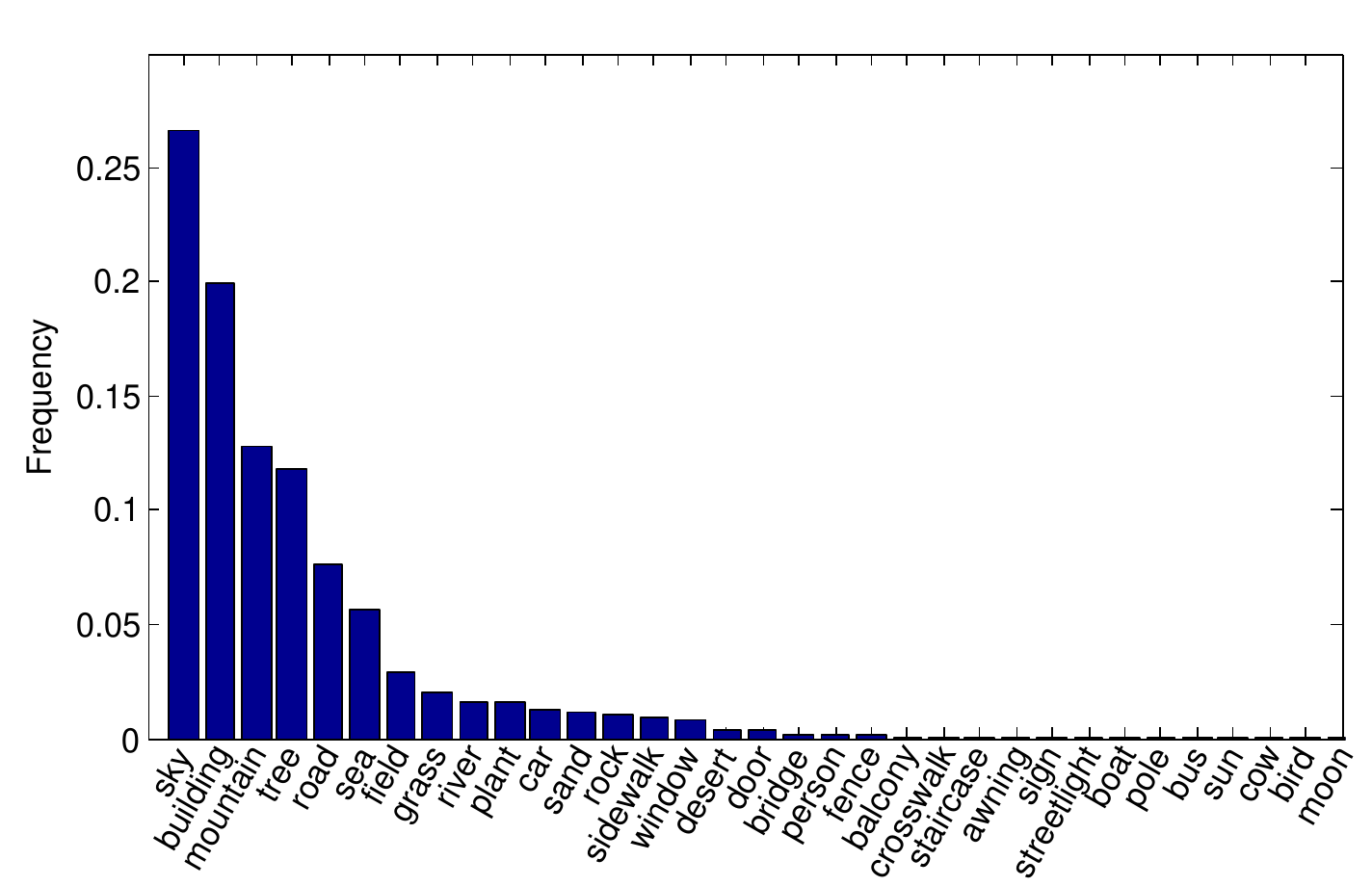}
  \hspace{-0.5em}
  \includegraphics[width=0.239\textwidth]{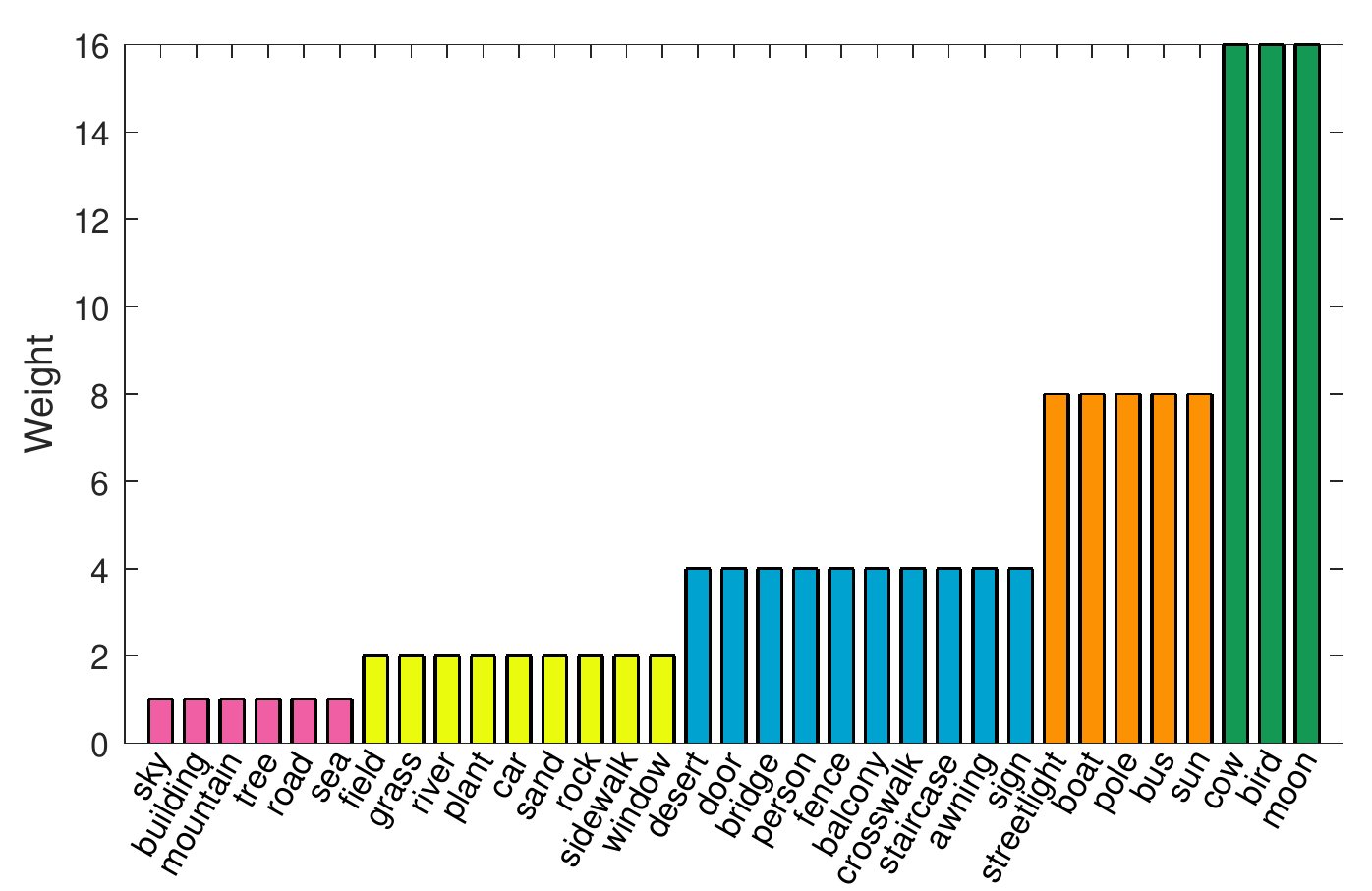}
\end{center}
\caption{Graphical visualization of the class frequencies (left) and weights (right) on the siftFlow datasets \cite{liu2009nonparametric}. The classes are sorted in the descending order based on their occurrence frequencies in training images.}
  \label{Figure:idf}
\end{figure}

In scene images, the class distribution is extremely imbalanced. Namely, very few classes account for large percentage of pixels in images. An example is demonstrated in Figure \ref{Figure:idf}. It's therefore common to put more attention to rare classes, in order to boost their recognition precisions.

In the patch-based CNN training, Farabet et al. \cite{farabet2013learning} and Shuai et al. \cite{shuai2015integrating} oversample the rare-class pixels to address this issue. It's however inapplicable to adopt this strategy in our network training, which is a complex structure learning problem.
Meanwhile, as the classes are distributed severely unequally in scene images, it's also problematic to weigh classes according to their inverse frequencies. As an example, the frequency ratio between the most frequent (sky) and the most rare class (moon) on the SiftFlow dataset is $3.5 \times 10^4$.  If the above class weighting criterion is adopted like in \cite{mostajabi2015feedforward}, the frequent classes will be under-attended.  Hence, we define the weighting function $\mathbf{w}$ as follows:
\begin{equation}
 \mathbf{w}_j = k^{\lceil log10(\eta / f_j) \rceil}
\end{equation}
where $\lceil \cdot \rceil$ is the integer ceiling operator, $f_j$ is the occurrence frequency of the class $j$, $\eta$ denotes the threshold that discriminates the rare classes. Specifically, a class is identified as rare if its frequency is smaller than $\eta$, otherwise, it is a frequent class. $k$ is a constant that controls the importance of rare classes ($k=2$ in our experiments). The proposed weighting function has the following properties: (1), it attends to rare classes by assigning them higher weights; (2), the degree of attention for rare classes grows exponentially based on their ratio magnitudes w.r.t the threshold $\eta$;
The following criterion is used to determine the value of $\eta$: the accumulated frequency of  all the non-rare classes is 85\%. We call it 85\%-15\% rule, and \cite{yang2014context} uses a similar rule.

\section{Experiments}
\label{Section:experiments}
We justify our method on three popular and challenging real-world scene image labeling benchmarks: SiftFlow \cite{liu2009nonparametric}, CamVid \cite{BrostowSFC:ECCV08} and Barcelona \cite{tighe2010superparsing}. Two types of scores are reported: the percentage of all correctly classified pixels (\textbf{Global}), and average per-class accuracy (\textbf{Class}).

\subsection{Baselines}
The convolution neural network (CNN), which jointly learn features and classifiers is used as our first baseline. In this case, the parameters are optimized to maximize the independent prediction accuracy for local patches. Another baseline is the network that shares the same architecture with our DAG-RNNs, while removes the recurrent connections. Mathematically, the $W_d$ and $b_d$ in Equation \ref{Equation:quad_dag_rnn} are fixed to $\zero$ . In this case, the DAG-recurrent neural network degenerates to an ensemble of four plain two-layer neural networks (CNN-ENN).
The performance disparity between the baselines and DAG-RNNs clearly illuminates the efficacy of our dependency modeling method.

\subsection{Implementation Details}

We use the following two networks to be the convolution layers in our experiments:
\begin{itemize}
  \item \textbf{CNN-65}: The network consists of five convolutional layers, the kernel sizes of which are $8\times8\times3\times64$, $6\times6\times64\times128$, $5\times 5 \times 128 \times 256$, $4 \times 4 \times 256 \times 256$ and $1 \times 1 \times 256 \times 64 $ respectively. Each of the first three convolutional layers are followed by a ReLU and non-overlapping $2 \times 2$ max pooling layer. The parameters of this network is learned from image patches ($65 \times 65$)
      of the target dataset only \textbf{(\underline{Setting 1})}.
  \item \textbf{VGG-conv5}: The network borrows its architecture and parameters from VGG-verydeep-16 net \cite{simonyan2014very}. In detail, we discard all the layers after the 5\textsuperscript{th} pooling layer to yield the desired convolution layer. The network is pre-trained on ImageNet dataset and fine-tuned on the target dataset. \cite{deng2009imagenet}
      \textbf{(\underline{Setting 2})}.
\end{itemize}

\begin{figure}
\newcommand{\InsertImage}[2]{
  \begin{subfigure}[t]{0.1\textwidth}
  \centering
  \includegraphics[width=\textwidth]{#1}
  \caption*{\scriptsize{{#2}}}
  \end{subfigure}
  }
\begin{center}
\InsertImage{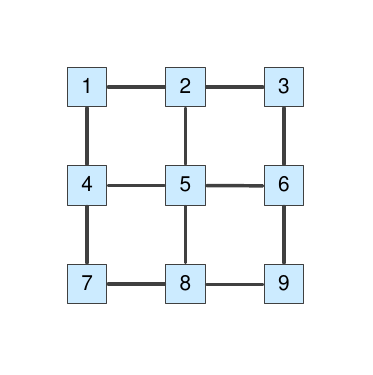}{UCG(4)}
\hspace{-5pt}
\InsertImage{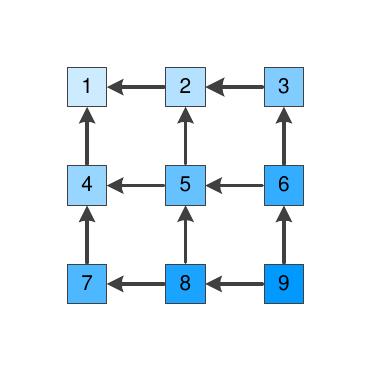}{DAG(4) $\mathcal{G}_{nw}^{4}$}
\begin{picture}(1,1)
\setlength{\unitlength}{0.11\textwidth}
\thicklines
\put(0,0){\line(0,1){1}}
\end{picture}
\InsertImage{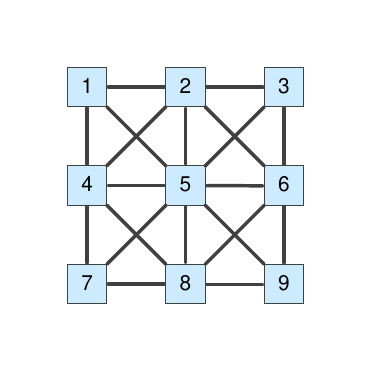}{UCG(8)}
\hspace{-5pt}
\InsertImage{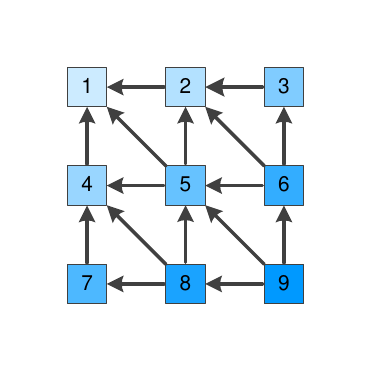}{DAG(8) $\mathcal{G}_{nw}^{8}$}
\end{center}
\caption{Two UCGs (with 4, 8 neighborhood system) and their induced DAGs in the northwestern (NW) direction.  }
\label{Figure:dags}
\end{figure}
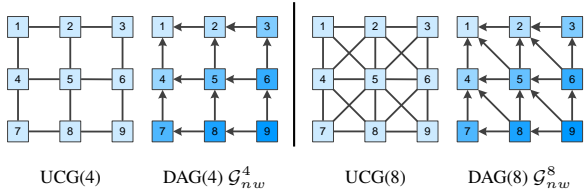

In DAG-RNNs, the adopted non-linear functions (refer to Equation \ref{Equation:dag_rnn}) are ReLU \cite{krizhevsky2012imagenet} for hidden neurons: $f(x) = max(0,x)$ and $softmax$ for output layer $g$. In practice, we apply the function $g$ after the deconvolution layer. The dimensionality of hidden layer $\hh$  is empirically set to 64 for CNN-65 and 128 for VGG-conv5 respectively.
\footnote{Based on our preliminary results, we didn't observe too much performance improvement by using larger h (e.g. 128 in CNN-65, and 256 in VGG-conv5) on the siftFlow dataset. In addition, the networks with larger capacity incur much heavier computation burdens.}
In our experiments, we consider two UCGs with 4 and 8 neighborhood systems. Their induced DAGs in the northwestern direction are shown in Figure \ref{Figure:dags}. In comparison with DAG(4), DAG(8) enables information to be propagated in shorter paths, which is critical to prevent the long-range information from vanishing. As exampled in Figure \ref{Figure:dags}, the length of propagation path from $v_9$ to $v_1$ in $\mathcal{G}_{nw}^8$ is halved to that in $\mathcal{G}_{nw}^4$ (4 $\rightarrow$ 2 steps).


The full network is trained by stochastic gradient descent with momentum. The parameters are updated after one image finishes its forward and backward passes. The learning rate is initialized to be $10^{-3}$, and decays exponentially with the rate of 0.9 after 10 epoch. The reported results are based on the model trained in 35 epoches.
We tune the parameters and diagnoses the network performance based on CNN-65. We also include the results of VGG-conv5 to see whether our proposed DAG-RNNs are beneficial for the highly discriminative representation from the state-of-the-art VGG-verydeep-16 net \cite{simonyan2014very}.

\subsection{SiftFlow Dataset}
\begin{table}[t]
\footnotesize
\begin{center}
\begin{tabular}{|l|cc|}
\hline
Methods & Global & Class \\
\hline
Byeon \emph{et al.} \cite{byeon2015scene} & 70.1\% & 22.6\% \\
Liu \emph{et al.}\cite{liu2009nonparametric} & 74.8\% & N/A \\
Farabet \emph{et al.} \cite{farabet2013learning} & 78.5\% & 29.4\% \\
Pinheiro \emph{et al.} \cite{pinheiro2014recurrent} & 77.7\% & 29.8\% \\
Tighe \emph{et al.}\cite{tighe2013finding} & 79.2\% & 39.2\% \\
Sharma \emph{et al.}\cite{sharma2014recursive} & 79.6\% & 33.6\% \\
Shuai \emph{et al.}\cite{shuai2015integrating} & \textbf{80.1}\% & 39.7\% \\
Yang \emph{et al.}\cite{yang2014context} & 79.8\% & \textbf{48.7}\% \\
\hline
CNN-65 & 76.1\% & 32.5\% \\
CNN-65-ENN & 76.1\% & 37.0\% \\
CNN-65-DAG-RNN(4) & 80.5\% & 42.6\% \\
CNN-65-DAG-RNN(8) & \textbf{81.1}\% & \textbf{48.2}\% \\
\hline
\hline
Long \emph{et al.}\cite{long2015fully} & 85.2\% & 51.7\% \\
\hline
VGG-conv5-ENN & 84.0\% & 48.8\% \\
VGG-conv5-DAG-RNN(8) & \textbf{85.3}\% & \textbf{55.7}\% \\
\hline
\end{tabular}
\end{center}
\caption{Quantitative performance of our method on the siftFlow dataset. The numbers (in brackets) following the DAG-RNN denote the neighborhood system of the UCG.}
\label{Table:result_SiftFlow}
\end{table}

The SiftFlow dataset has 2688 images generally captured from 8 typical outdoor scenes. Every image has $256 \times 256$ pixels, which belong to one of the 33 semantic classes. We adopt the training/testing split protocol (2488/200 images) provided by \cite{liu2009nonparametric} to perform our experiments.  Following the 85\%-15\% criterion, the class frequency threshold $\eta = 0.05$. Statistically, out of 33 classes, 27 of them are regarded as infrequent class. The graphical visualization of the weights for different classes are depicted in Figure \ref{Figure:idf}.

The quantitative results are listed in Table \ref{Table:result_SiftFlow}, within which the upper part presents the performance of methods under setting 1.
Our baseline CNN-65 achieves very promising results, which proves the effectiveness of the convolution layer. We also notice that results of CNN-65 fall behind CNN-65-ENN on the average class accuracy. This phenomenon is also observed on the CamVid and Barcelona benchmarks, as shown by Table \ref{Table:result_CamVid} and \ref{Table:result_Barcelona} respectively.
This result indicates that the proposed class weighting function significantly boosts the recognition accuracy for rare classes.
By adding DAG-RNN(8), our full network reaches 81.1\% (48.2\%) on the global (class) accuracy
\footnote{If we disassemble the full labeling network to two disjoint parts - CNN-65 and DAG-RNN(8), and they are optimized independently, the corresponding accuracies are 80.1\% and 42.7\%. The performance discrepancy indicates the importance of the joint optimization for the full network.}
, which outperforms the baseline (CNN-65-ENN) by 5\% (11.2\%). Meanwhile, we observe promising accuracy gain (global: 0.6\% / class: 5.6\% ) by switching DAG-RNN (4) to DAG-RNN (8), in which we believe that long-range dependencies are better captured as information propagation paths in DAG(8) are shorter than those in DAG(4). Such performance benefits can be observed consistently on the CamVid (0.5\% / 2.0\%) and Barcelona (1.1\% / 1.6\%) datasets, as evidenced in Table \ref{Table:result_CamVid} and \ref{Table:result_Barcelona} respectively. Moreover, in comparison with other representation learning nets, which are fed with much richer contextual input (133$\times$133 patch in \cite{pinheiro2014recurrent}, 3-scale 46$\times$46 patches in \cite{farabet2013learning}), our DAG-RNNs outperform theirs by a large margin.  Importantly, our results match the state-of-the-art under this setting.

Furthermore, we initialize our convolution layers with VGG-verydeep-16 \cite{simonyan2014very}, which has been proven to be the state-of-the-art feature extractor. The quantitative results under setting 2 are listed in the lower body of Table \ref{Table:result_SiftFlow}. Our baseline VGG-conv5-ENN surpasses the best performance of methods under setting 1. This result indicates the significance of large-scale data in deep neural network training. Interestingly, our DAG-RNN(8) is still able to further improve the discriminative power of local features by modeling their dependencies, thereby leading to a phenomenal (6.9\%) average class accuracy boost. Note that Fully Convolution Networks (FCNs) \cite{long2015fully} uses activations (feature maps) from multiple convolution layers, whereas our VGG-conv5-ENN only use feature maps from conv5 layer. Hence, there is a slight performance gap between our VGG-conv5-ENN and FCNs. Nonetheless, our VGG-conv5-DAG-RNN(8) still performs comparably with FCNs on global accuracy, and significantly outperforms it on the class accuracy. Importantly, our full labeling network also achieves new state-of-the-art performance under this setting. The detailed per-class accuracy is listed in Table \ref{Table:SiftFlow_class_accuracy}.

\subsection{CamVid Dataset}
\begin{table}[t]
\footnotesize
\begin{center}
\begin{tabular}{|l|cc|}
\hline
Methods & Global & Class \\
\hline
Tighe \emph{et al.}\cite{tighe2010superparsing} & 78.6\% & 43.8\% \\
Sturgess \emph{et al.}\cite{sturgess2009combining} & 83.8\% & 59.2\%\\
Zhang \emph{et al.}\cite{zhang2010semantic} & 82.1\% & 55.4\% \\
Bulo \emph{et al.}\cite{bulo2014neural} & 82.1\% & 56.1\% \\
Ladicky \emph{et al.}\cite{ladicky2010and} & 83.8\% & \textbf{62.5}\% \\
Tighe \emph{et al.} \cite{tighe2013finding} & \textbf{83.9}\% & \textbf{62.5}\% \\
\hline
CNN-65  & 84.3\% & 53.2\% \\
CNN-65-ENN & 84.1\% & 58.1\% \\
CNN-65-DAG-RNN(4) & 88.2\% & 66.3\% \\
CNN-65-DAG-RNN(8) & \textbf{88.7}\% & \textbf{68.3}\% \\
\hline
\hline
VGG-conv5-ENN & 91.0\% & 76.5\% \\
VGG-conv5-DAG-RNN (8) & \textbf{91.6}\% & \textbf{78.1}\% \\
\hline
\end{tabular}
\end{center}
\caption{Quantitative performance of our method on the CamVid dataset. }
\label{Table:result_CamVid}
\end{table}
The CamVid dataset \cite{BrostowSFC:ECCV08} contains 701 high-resolution images ($960 \times 720$ pixels) from 4 driving videos at daytime and dusk (3 daytime and 1 dusk video sequence). Images are densely labelled with 32 semantic classes. We follow the usual split protocol \cite{sturgess2009combining}\cite{tighe2013finding} (468/233) to obtain training/testing images. Similar to other works \cite{BrostowSFC:ECCV08}\cite{bulo2014neural}\cite{sturgess2009combining}\cite{tighe2013finding}, we only report results on the most common $11$ categories. According to the 85\%-15\% rule, 4 classes are identified as rare, and $\eta$ is 0.1.

The quantitative results are given in Table \ref{Table:result_CamVid}. Our baseline networks (CNN-65, CNN-65-ENN) achieve very competitive results. By explicitly modeling contextual dependencies among image units, our CNN-65-DAG-RNN(8) brings phenomenal performance benefit (4.6\% and 10.2\% for the global and class accuracy respectively). Moreover, in comparison with state-of-the-art methods \cite{bulo2014neural}\cite{ladicky2010and}\cite{sturgess2009combining} \cite{tighe2013finding},  our CNN-65-DAG-RNN(8) outperforms theirs by a large margin (4.8\% / 5.8\%), demonstrating the profitability of adopting high-level features learned from CNN and context modeling with our DAG-RNNs.
Furthermore, the VGG-conv5-ENN alone performs excellently. Even though the performance starts saturating, our DAG-RNN(8) is able to consistently improve the labeling results.



\subsection{Barcelona Dataset}
The barcelona dataset \cite{tighe2010superparsing} consists of 14871 training
and 279 testing  images. The size of the images varies across different instances, and each pixel is labelled as one of the 170 semantic classes. The training images range from indoor to outdoor scenes, whereas the testing images are only captured from the barcelona street scene. These issues pose Barcelona as a very challenging dataset. Based on the 85\%-15\% rule, 147 classes are identified as rare classes, and the class frequency threshold $\eta$ is 0.005.

Table \ref{Table:result_Barcelona} presents the quantitative results. From which, we clearly observe that our baseline networks (CNN-65 and CNN-65-ENN) achieve very competitive results, which has already matched the state-of-the-art results. The introduction of DAG-RNN(8) leads to promising performance improvement, therefore the full labeling network clinches the new state-of-the-art under setting 1. More importantly, under setting 2, even though the VGG-conv5-ENN is extraordinarily competitive, the DAG-RNN(8) is still able to enhance its labeling performance significantly.

\begin{table}[t]
\footnotesize
\begin{center}
\begin{tabular}{|l|cc|}
\hline
Methods & Global & Class \\
\hline
Tighe \emph{et al.}\cite{tighe2010superparsing} & 66.9\% & 7.6\% \\
Farabet \emph{et al.} \cite{farabet2013learning} & 46.4\% & \textbf{12.5}\% \\
Farabet \emph{et al.} \cite{farabet2013learning} & \textbf{67.8}\% & 9.5\%\\
\hline
CNN-65 & 69.0\% & 10.5\% \\
CNN-65-ENN & 69.0\%  & 11.0\% \\
CNN-65-DAG-RNN(4) & 71.3\% & 12.9\% \\
CNN-65-DAG-RNN(8) & \textbf{72.4}\% & \textbf{14.5}\% \\
\hline
\hline
VGG-conv5-ENN & 73.3\% & 21.1\% \\
VGG-conv5-DAG-RNN(8) & \textbf{74.6}\% & \textbf{24.6}\% \\
\hline
\end{tabular}
\end{center}
\caption{Quantitative performance of our method on the Barcelona dataset.}
\label{Table:result_Barcelona}
\end{table}
\subsection{Effects of DAG-RNNs to Per-class Accuracy}


\begin{table*}[t]
\setlength{\tabcolsep}{2.25pt}
\tiny
\begin{center}
\begin{tabular}{|l|cccccc|cccccccccccccccccccccccc|cc|}
\hline
& \rotatebox{90}{sky} & \rotatebox{90}{building} & \rotatebox{90}{tree} & \rotatebox{90}{mountain} & \rotatebox{90}{road} & \rotatebox{90}{sea} & \rotatebox{90}{field} & \rotatebox{90}{car} & \rotatebox{90}{sand} & \rotatebox{90}{river} & \rotatebox{90}{plant} & \rotatebox{90}{grass} & \rotatebox{90}{window} & \rotatebox{90}{sidewalk} & \rotatebox{90}{rock} & \rotatebox{90}{bridge} & \rotatebox{90}{door} & \rotatebox{90}{fence} & \rotatebox{90}{person} & \rotatebox{90}{staircase} & \rotatebox{90}{awning} & \rotatebox{90}{sign} & \rotatebox{90}{boat} & \rotatebox{90}{crosswalk} & \rotatebox{90}{pole} & \rotatebox{90}{bus} & \rotatebox{90}{balcony} &\rotatebox{90}{streetlight} & \rotatebox{90}{sun} & \rotatebox{90}{bird} &
\rotatebox{90}{global} & \rotatebox{90}{class}\\
\hline
\tiny{Frequency} & 27.1 & 20.2 & 12.6 & 12.4 & 6.93 & 5.61 & 3.64 & 1.64 & 1.41 & 1.37 & 1.33 & 1.22 & 1.07 & 0.89 & 0.85 & 0.36 & 0.26 & 0.24 & 0.23 & 0.18 & 0.11 & 0.11 & 0.06 & 0.05 & 0.04 & 0.03 & 0.03 & 0.02 & 0.01 & 0.004 & - & -\\
\hline
\hline
\tiny{CNN-65-ENN} & 94.1 & 84.9 & 77.6 & 74.2 & 80.9 & 61.1 & 30.7 & 71.0 & 27.7 & 34.7 & 21.8 & 63.5 & 27.8 & 47.0 & 21.0 & 8.8 & 35.7 & 33.7 & 29.3 & 6.8 & 0.37 & 16.3 & 1.4 & 48.1 & 0 & 0.43 & 34.5 & 6.0 & 71.4 & 0 & 76.1 & 37.0 \\
\tiny{CNN-65-DAG-RNN(8)} & \textbf{95.9} & \textbf{87.3} & \textbf{82.5} & \textbf{77.9} & \textbf{85.8} & \textbf{70.2} & \textbf{43.5} & \textbf{80.1} & \textbf{52.9} & \textbf{65.4} & \textbf{37.2} & 57.8 & \textbf{40.4} & \textbf{59.1} & \textbf{27.6} & \textbf{31.4} & \textbf{51.8} & \textbf{38.0} & \textbf{35.6} & \textbf{50.9} & \textbf{7.8} & \textbf{31.0} & \textbf{5.14} & \textbf{82.8} & 0 & 0 & \textbf{54.9} & \textbf{8.59} & \textbf{85.5} & 0 &\textbf{81.1} & \textbf{48.2} \\
\hline
\hline
\tiny{VGG-conv5-ENN} & 96.0 & 91.1 & 84.4 & 82.9 & 90.4 & 83.5 & 48.8 & 77.5 & 63.5 & 57.6 & 32.6 & 60.2 & 34.9 & 66.0 & 25.8 & 20.0 & 51.9 & 44.0 & 38.6 & 45.9 & 26.5 & 33.7 & 14.9 & 50.2 & 1.1 & 0 & 32.7 & 9.1 & 99.9 & 0 & 84.0 & 48.8 \\
{VGG-conv5-DAG-RNN(8)} & \textbf{96.3} & 90.8 & 82.1 & \textbf{85.1} & 89.2 & \textbf{84.8} & \textbf{55.4} & \textbf{84.2} & \textbf{67.9} & \textbf{75.3} & \textbf{51.5} & \textbf{64.8} & \textbf{45.2} & 63.5 & \textbf{45.7} & \textbf{37.3} & \textbf{56.8} & \textbf{44.7} & 36.2 & \textbf{58.7} & 18.3 & \textbf{40.0} & \textbf{63.3} & \textbf{65.2} & \textbf{18.4} & \textbf{1.4} & \textbf{45.8} & 5.4 & 97.9 & 0 & \textbf{85.3} & \textbf{55.7} \\
\hline
\end{tabular}
\end{center}
\caption{Per-class accuracy comparison on the SiftFlow dataset. All the numbers are displayed in the percentage scale. The statistics for class frequency is obtained in test images. For reading convenience, the frequent and rare classes are placed in the same block.}
\label{Table:SiftFlow_class_accuracy}
\end{table*}

In this section, we investigate the effects of our DAG-RNNs for each class. The detailed per-class accuracy for the SiftFlow dataset is listed in Table \ref{Table:SiftFlow_class_accuracy}.
Under setting 1, we find that the contextual information encoded through our DAG-RNN(8) is beneficial for almost all classes.
In this case, the local representations from CNN-65 are not strong, so their discriminative power can be greatly enhanced by modeling their dependencies.
In line with it, we observe remarkable performance benefit (+11.2\%) for almost all classes.
Under setting 2, the VGG-conv5 net is pre-trained on the ImageNet dataset \cite{deng2009imagenet}, and it recognizes most classes excellently.
Even though the local representations are highly discriminative in this situation, our DAG-RNN(8) further tremendously improves their representative power for rare classes.
Statistically, we observe a phenomenal 8.6\% accuracy gain for rare classes.
Under both settings, modeling the dependencies among local features enables the classification to be contextual aware.  Therefore, the local ambiguities are mitigated to a large extent.
However, we fail to observe commensurate accuracy improvements for extremely-small-size and rare 'object' classes (e.g. bird and bus), we conjecture that the weak local information may have been overwhelmed by context (e.g. a small bird is swallowed by the broad sky in Figure \ref{Figure:intro}).




\begin{figure*}
\newcommand{\InsertImageS}[1]{
\begin{subfigure}[t]{0.11\textwidth}
\includegraphics[width=\textwidth]{#1}
\end{subfigure}
\hspace{-0.01\textwidth}
}
\newcommand{\InsertImageD}[3]{
\begin{subfigure}[t]{0.11\textwidth}
\captionsetup{
justification=raggedleft,singlelinecheck=false, position=top
}
\centering
\includegraphics[width=\textwidth]{#1}
\caption*{\footnotesize{{#2 (#3)}}}
\end{subfigure}
\hspace{-0.01\textwidth}
}
\begin{center}
\InsertImageS{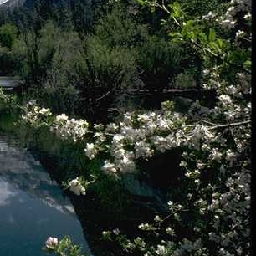}
\InsertImageD{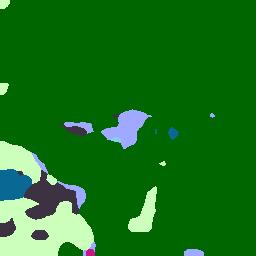}{90.8\%}{77.6\%}
\InsertImageD{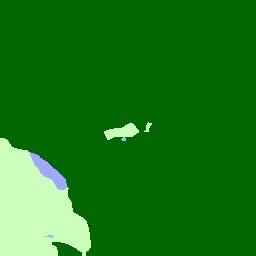}{97.5\%}{95.3\%}
\InsertImageS{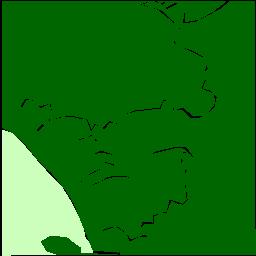}
\hspace{0.03\textwidth}
\InsertImageS{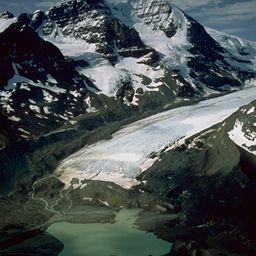}
\InsertImageD{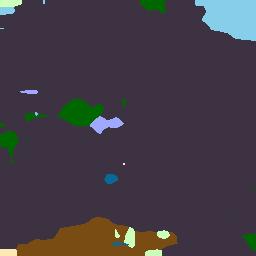}{86.9\%}{56.1\%}
\InsertImageD{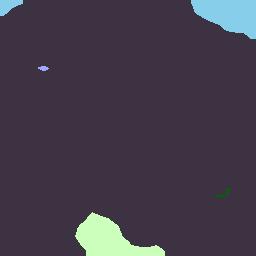}{94.8\%}{72.7\%}
\InsertImageS{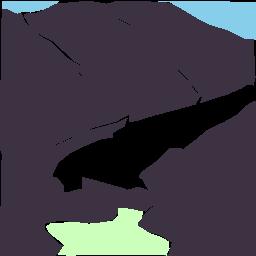}\\
\vspace{-6pt}

\InsertImageS{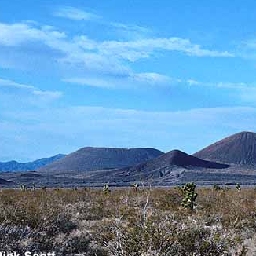}
\InsertImageD{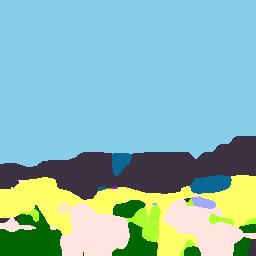}{75.5\%}{69.7\%}
\InsertImageD{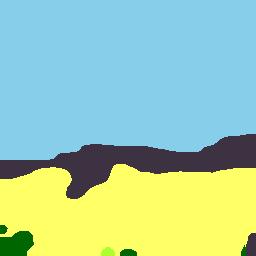}{94.9\%}{91.5\%}
\InsertImageS{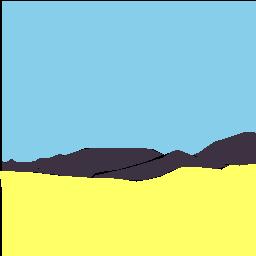}
\hspace{0.03\textwidth}
\InsertImageS{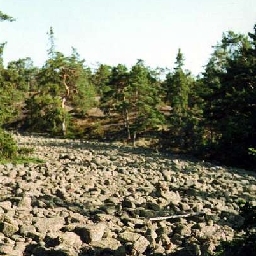}
\InsertImageD{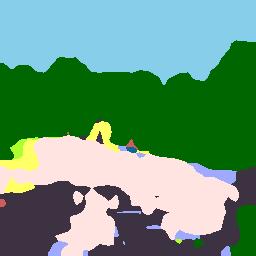}{55.9\%}{65.2\%}
\InsertImageD{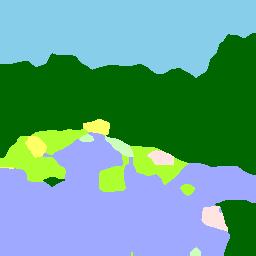}{85.4\%}{87.7\%}
\InsertImageS{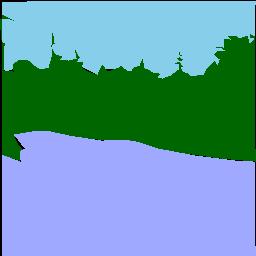}\\
\vspace{-6pt}

\InsertImageS{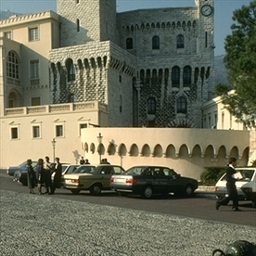}
\InsertImageD{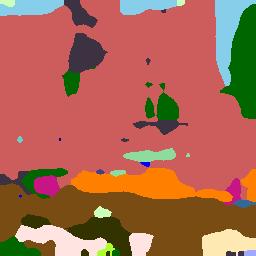}{74.8\%}{50.9\%}
\InsertImageD{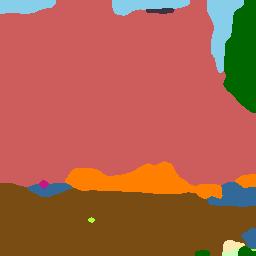}{87.2\%}{52.4\%}
\InsertImageS{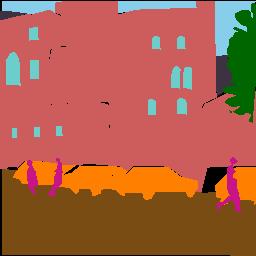}
\hspace{0.03\textwidth}
\InsertImageS{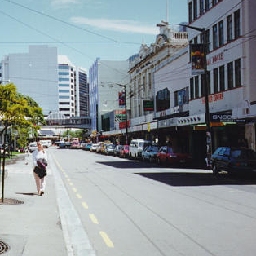}
\InsertImageD{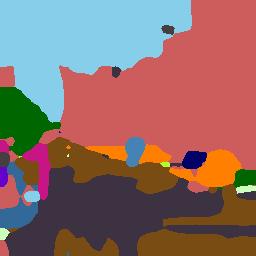}{64.0\%}{57.3\%}
\InsertImageD{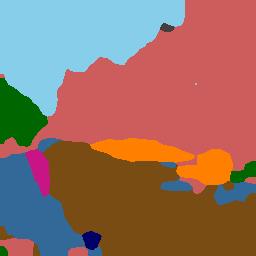}{83.2\%}{72.8\%}
\InsertImageS{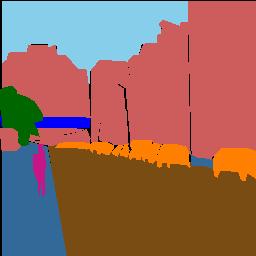}\\
\vspace{-6pt}

\InsertImageS{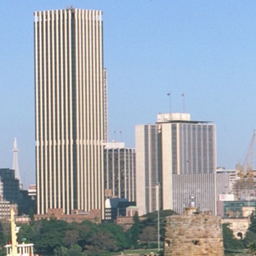}
\InsertImageD{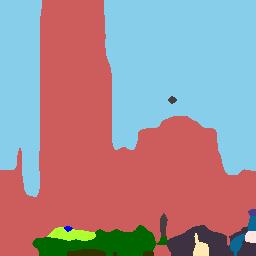}{87.9\%}{75.6\%}
\InsertImageD{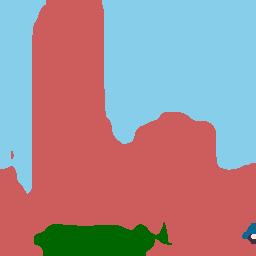}{94.3\%}{85.9\%}
\InsertImageS{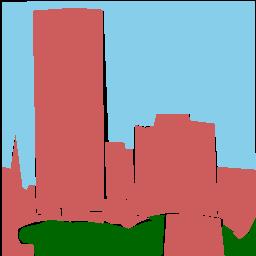}
\hspace{0.03\textwidth}
\InsertImageS{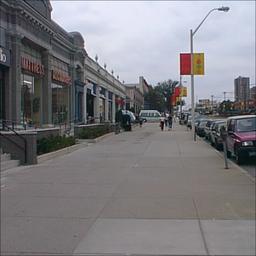}
\InsertImageD{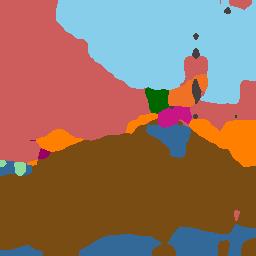}{55.1\%}{50.8\%}
\InsertImageD{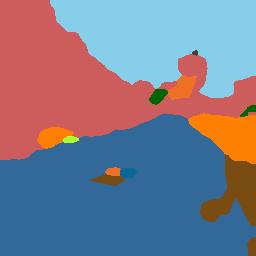}{87.2\%}{42.3\%}
\InsertImageS{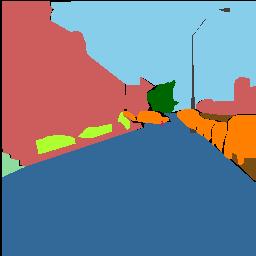}\\

\vspace{-6pt}
\includegraphics[width=0.5\textwidth]{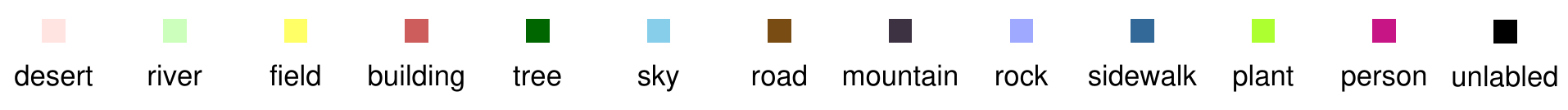}

\end{center}
\caption{Qualitative labeling results (best viewed in color). We show input images, local prediction maps (CNN-65-ENN), contextual labeling maps (CNN-65-DAG-RNN(8)) and their ground truth respectively. The numbers outside and inside the parentheses are global and class accuracy respectively. }
\label{Figure:results_final}
\end{figure*}

\subsection{Discussion of Modeled Dependency}
\vspace{-3pt}
We show a number of qualitative labeling results in Figure \ref{Figure:results_final}.
By looking into them, we can have some interesting observations. The DAG-RNNs are capable of (1), enforcing  local consistency: neighborhood pixels are likely to be assigned to the same labels. In Figure \ref{Figure:results_final}, the left-panel examples show that confusing regions are smoothed by using our DAG-RNNs. (2), ensuring semantic coherence: the pixels that are spatially far away  are usually given labels that could co-occur in a meaningful scene. For example, the `desert' and `mountain' classes are usually not seen together with `trees' in a `open country' scene, so they are corrected to `stone' in the second example of the right panel.  More examples of this kind are shown in the right panel. These results illuminate that short-range and long-range contextual dependencies may have been captured by our DAG-RNNs.


\section{Conclusion}
\label{Section:conclusion}
\vspace{-5pt}
In this paper,
we propose DAG-RNNs to process DAGs-structured data, where the interactions among local features are considered in a graphical structure. Our DAG-RNNs are capable of encoding the abstract gist of images into local representations, which tremendously enhance their discriminative power. Furthermore, we propose a novel class weighting function to address the imbalanced class distribution issue, and it is experimentally proved to be effective towards the recognition enhancement for rare classes. Integrating with the convolution and deconvolution layers, our DAG-RNNs achieve state-of-the-art results on three challenging scene labeling benchmarks. We also demonstrate that useful long-range contextual dependencies are captured by our DAG-RNNs, which is helpful for generating smooth and semantically sensible labeling maps in practice.


{\small
\bibliographystyle{ieee}
\bibliography{cvpr-2016}
}

\end{document}